%% file: paper.tex
\documentclass[preprint]{article}
\usepackage[numbers]{natbib}
\usepackage{amsmath}
\usepackage{amsfonts}
\usepackage{hyperref}
\usepackage{breakurl}
\usepackage{booktabs}
\usepackage[table]{xcolor}
\usepackage[nonatbib]{neurips_2024}

\title{Reinforcement Learning without Human Feedback for Last Mile Fine-Tuning of Large Language Models}
\author{{\fontsize{11}{11}\selectfont Alec Solway}\\[3pt]Two Six Technologies}

\begin{document}
\maketitle

\begin{abstract}
Reinforcement learning is used to align language models with human preference signals after first pre-training the model to predict the next token of text within a large corpus using likelihood maximization. Before being deployed in a specific domain, models are often further fine-tuned on task specific data. Since human preferences are often unavailable for the last step, it is performed using likelihood maximization as that is the typical default method. However, reinforcement learning has other advantages besides facilitating alignment to a human derived reward function. For one, whereas likelihood maximization is a form of imitation learning in which the model is trained on what to do under ideal conditions, reinforcement learning is not limited to demonstrating actions just for optimally reached states and trains a model what to do under a range of scenarios as it explores the policy space. In addition, it also trains a model what not to do, suppressing competitive but poor actions. This work develops a framework for last-mile fine-tuning using reinforcement learning and tests whether it garners performance gains. The experiments center on abstractive summarization, but the framework is general and broadly applicable. Use of the procedure produced significantly better results than likelihood maximization when comparing raw predictions. For the specific data tested, the gap could be bridged by employing post-processing of the maximum likelihood outputs. Nonetheless, the framework offers a new avenue for model optimization in situations where post-processing may be less straightforward or effective, and it can be extended to include more complex classes of undesirable outputs to penalize and train against, such as hallucinations.
\end{abstract}

\section{Introduction}
Reinforcement learning from human feedback (RLHF) is a standard key ingredient for fine-tuning large language models to follow instructions and produce fluent dialogue~\cite{googleGoogleGeminiAI2024,openaiGPT4TechnicalReport2023,ouyangTrainingLanguageModels2022a,touvron2023llama}. In the typical setup, a foundation model is first trained via maximum likelihood (equivalently, cross entropy minimization) to predict the next token of text conditional on a window of prior tokens. A similar setup is then used in a first fine-tuning step to train the model on a target set of tasks based on ground-truth input-output pairs. Following the first stage of fine-tuning, the model's output is stochastically sampled in response to a series of further prompts, and the outputs are ranked by human annotators. The rankings are used as the basis for a second stage of fine-tuning using reinforcement learning. This stage may be iterated several times to further improve the model's performance.

Reinforcement learning has a number of standard components. In this context, reward is derived from human rankings either by training a model on the ranking data or by referencing them directly, the state space consists of the input/output tokens produced so far, the actions are the tokens themselves, and the state transition dynamics are trivially defined as the result of appending an additional token. With these ingredients in place, a policy network is trained using either proximal policy optimization (PPO)~\cite{schulman2017proximal} if a reward model is learned or directly using direct preference optimization~\cite{rafailov2024direct}.

After training a general version of the model to perform as well as possible across a range of tasks, the model may yet further be fine-tuned to specialize in a particular task for a specific use case. This final fine-tuning is typically done via maximum likelihood, not reinforcement learning, as in the very first step described above. RLHF produces state-of-the-art results for several reasons. First, human rankings provide direct feedback on how well a model accomplishes a particular objective, such as engaging in fluid and useful conversation. Second, training examples include not only what outputs are good but also what outputs are bad, that is, lower ranked or not chosen. Third, although direct preference optimization has other advantages, when PPO is employed, a hybrid model-based/model-free RL scheme is used to simulate a range of additional training examples which likely help the model learn what to do when reaching suboptimal states. In contrast, likelihood maximization only tells a model what should be done under optimal conditions, not what it should not do or what second best performance would look like if perfect recall is not possible.

While human rankings of outputs are clearly valuable for guiding model training, they can be costly and time consuming to obtain and are not always available, which is why the last step of domain specific fine-tuning often does not incorporate them. However, the second and third advantages motivate the question of whether incorporating reinforcement learning into last-mile fine-tuning based on data without human rankings can garner performance gains. Testing this formed the focus of the current work. The same tools used for traditional RLHF were imported with minimal modification, with PPO-based optimization used because the data were not rank-based. A reward model was trained on a combination of the ground-truth outputs and simple examples of negative outputs. Similar to RLHF, a reward model of this type in essence represents a similarity metric between the output presented to it and the best output to produce as determined by generalizing across the ground-truth data. PPO was used in combination with a hybrid model-based/model-free RL scheme, also similar to RLHF, to simulate different training examples. Besides the reward model, which is trained on static data, the state transition dynamics are naturally and trivially defined as noted above. PPO-based optimization runs a series of simulations using the state transition and reward models together with the current policy to produce a range of examples which in turn are used to further refine the policy network to perform well under different scenarios (states). The experiments centered on abstractive summarization, but the general framework is broadly applicable to any task. 

\section{Related work}
Reinforcement learning had been applied to the topic of summarization prior to the modern LLM era. A number of papers approached the problem by directly maximizing ROUGE or a variant as part of the criterion (reward) being optimized. For example, \citet{pasunuruMultiRewardReinforcedSummarization2018a} used a combination of standard ROUGE, a modified ROUGE augmented to upweight salient words detected with a saliency classifier, and an entailment measure output by a separate entailment model. These criteria were optimized together by a policy network with an LSTM architecture. \citet{paulus2018a} combined standard next-token likelihood and ROUGE and an LSTM architecture with specific attention mechanisms aimed at reducing repetition in the output. \citet{li2018actor} used a gated recurrent unit (GRU) architecture with reward based on a combination of the next-token likelihood and classifier output distinguishing ground-truth and generated summaries. \citet{chenFastAbstractiveSummarization2018} employed convolution at the sentence level combined with an LSTM architecture across sentences in a model which first extracts salient sentences and then summarizes them, with ROUGE-L playing the role of reward. \citet{wuLearningExtractCoherent2018} used a combined convolution/GRU architecture and a combination of ROUGE and coherence for reward, with the latter learned by a subnetwork trained on positive (adjacent) and negative (random) sentence pairs in the training data. \citet{keneshlooDeepTransferReinforcement2019} used ROUGE and an LSTM architecture with a specific focus on transfer learning, training the model on mixed samples from more than one dataset. Combining RL-based approaches with the richer semantic representations of a pre-trained language model, \citet{baeSummaryLevelTraining2019} and \citet{wangTextAbstractionSummary2019} operated over BERT embeddings instead of starting with raw inputs while still using ROUGE for reward. Moving away from ROUGE, \citet{liDeepReinforcementLearning2019} used BERTScore to reward semantic similarity between the target and output instead of training the model to capture the target exactly. The use of reinforcement learning in these contexts was often motivated by its ability to optimize a non-differentiable metric like ROUGE.

There is of course a large literature on summarization without appealing to reinforcement learning. The present focus is specifically on the ability to leverage pre-trained large language models and not on specialized architectures. \citet{liuTextSummarizationPretrained2019} used sentence and document level representations built on BERT with a summary network trained using maximum likelihood. PEGASUS~\cite{pmlr-v119-zhang20ae} is a larger Transformer based language model pre-trained with summarization specifically in mind. During training, important sentences rather than random words were masked and recovered, paralleling extractive summarization. T5 is a Transformer based encoder-decoder model built based on the results of a systematic evaluation of how different choices during the modeling process impact performance. It achieved state-of-the-art performance at the time on a collection of tasks which included summarization~\cite{raffelExploringLimitsTransfer2020}. Both PEGASUS and T5 were trained using a likelihood-based objective. As noted in the Introduction, many modern models utilize a hybrid maximum likelihood/RL training procedure, using reinforcement learning to align the model with human preferences. \citet{stiennonLearningSummarizeHuman2020} is a prominent early example of this idea in which the authors fine-tuned GPT3 for summarization specifically. Llama 2, Gemini and modern versions of GPT all use this approach~\cite{googleGoogleGeminiAI2024,openaiGPT4TechnicalReport2023,ouyangTrainingLanguageModels2022a,touvron2023llama}.

The hybrid model-based/model-free RL scheme employed here and in PPO-based versions of RLHF is a form of data augmentation. Data augmentation is common in the vision domain, and while less frequent, has also been applied to text data~\cite{bayerSurveyDataAugmentation2023,fengSurveyDataAugmentation2021,weiEDAEasyData2019a}. It has largely been used to make models robust to noise and variability in the input, especially for small datasets. Here, augmentation is instead of the output, aimed at helping the model both avoid poor outputs and perform well in a range of different states.

\section{Models}
\subsection{General modeling procedures} A 4bit GPTQ quantized version of Llama2-7B-Chat was used as the base model for all others, including the reward, value, and policy networks, and the maximum likelihood fine-tuned control model. Training was performed using low rank adaptation~\cite{hu2021lora} with $r=16$ and $\alpha=16$ for the reward network and $r=32$ and $\alpha=16$ for the other models. The reward and value networks had a different final linear layer that projected the model's state to a scalar instead of to token logits as in the language model. Optimization of all models was performed using AdamW with a cosine learning rate schedule starting at $1\text{e-}5$; training consisted of one epoch. The batch size was 14 for training the reward models and 7 for PPO and maximum likelihood optimization.

\subsection{Reward modeling} Training a reward model requires not just positive examples but also negative examples that represent undesirable outputs. In the current work, five basic categories of negative examples were generated based on the positive examples in the original dataset:
\begin{enumerate}
    \item Completely random tokens for the outputs paired with existing inputs.
    \item Existing input and output sequences randomly re-paired, representing coherent but irrelevant outputs.
    \item Words (for simplicity, whitespace delineated entities) from an existing output sequence randomly shuffled and paired with the same original input. The output is a bag of the correct words without coherence.
    \item Outputs that begin correctly but end with repetitive sequences, which is a pattern commonly produced by models fit via maximum likelihood.
    \item Outputs which repeat the input sequence, also a common failure mode for maximum likelihood optimized models.
\end{enumerate}

Additional categories representing more complex errors, such as hallucinations or poor summarizations, can also be added as noted in the Discussion. The size of each class was equal to the size of the original dataset. To account for the class imbalance when training the reward network, each negative datum was weighted by a factor of $1/5$. Each token in the output of a positive example was assigned reward $+1$ and each token in the output of a negative example was assigned reward $0$. This left a trail of breadcrumbs for the agent to follow during policy optimization. More complex categories of negative examples may require reward at the sequence rather than token level, which may be harder to optimize. The reward model was trained using a squared loss function to predict a scalar conditional on the input tokens and all preceding output tokens. During policy learning, the reward predicted by the model was additionally combined with a length penalty of $-2.5$ for each token produced beyond the ground-truth length.

\subsection{Policy and value models} In the current work one epoch of standard proximal policy optimization~\cite{schulman2017proximal} was applied for each outer training batch with all trajectories combined in a single mini-batch (i.e. there was a single gradient update for each outer batch), although this represents an avenue for further refinement. For half of the outer loop batches, outputs were randomly sampled from the policy network, and the remaining half used ground-truth outputs. For completeness, the policy loss was:
\begin{align}
    w_t(\theta) &= \frac{\pi_\theta(a_t|s_t)}{\pi_{\theta_{\text{old}}}(a_t|s_t)},  \\
    A_t   &= \sum_{t'=t}^{T-1} (\gamma\lambda)^{t'-t}(r_{t'+1} + \gamma V(s_{t'+1}) - V(s_{t'})), \\
    L_{PPO}     &= \mathbb{E}_t \left[ \min(w_t(\theta)A_t, clip(w_t(\theta), 1-\epsilon,1+\epsilon)A_t) \right],
\end{align}
where $r_t$ is the reward from the reward model plus the length penalty, $T$ is the length of the output sequence, $\lambda=0.95$ and $\gamma=0.99999$. Note that with a single update the clipping is redundant. For simplicity, the loss did not include an entropy term or the KL divergence with the original model. However, the use of LoRA acts as an implicit prior, similar to incorporating an explicit KL divergence term. A squared loss was used for value with the target defined as:
\begin{align}
    V(s_t)^{target} = \sum_{t'=t}^{T-1} (\gamma)^{t'-t}r_{t'+1} \quad \forall t \in [0, T-1]
\end{align}
and $V(s_T)^{target} = 0$.

\subsection{Control models} Two classes of models were used as controls for the experiments: the base model without any fine-tuning and the base model fine-tuned using maximum likelihood. 

\section{Experiments}

\subsection{Metrics} ROUGE and BLEURT~\cite{puLearningCompactMetrics2021} were used to evaluate the models. ROUGE is a simple well-known metric that captures n-gram overlap, while BLEURT is a model-based measure built on BERT embeddings to predict human evaluations. ROUGE is confounded when the lengths of the predicted and reference summaries are different: a longer predicted summary has a larger chance of increasing the numerator for $ROUGE_{recall}$ while the denominator is fixed. Likewise, a longer reference summary has a larger chance of increasing the numerator for $ROUGE_{precision}$. To account for this, length adjusted (la-) versions of ROUGE were computed in addition to the standard versions. For recall, the standard measure was multiplied by $ng/np$ when $np>ng$ and by $1$ otherwise, where $ng$ is the number of whitespace delineated entities in the ground-truth summary and $np$ is the number of such entities in the prediction. For precision, the standard measure was multiplied by $np/ng$ when $np<ng$ and by $1$ otherwise. 

\subsection{Datasets} Performance was tested using two datasets: samsum~\cite{gliwaSAMSumCorpusHumanannotated2019}, consisting of short conversations, and xsum~\cite{narayanDonGiveMe2018}, consisting of news articles and single sentence summaries. In order to fit the models in a reasonable amount of time on the available hardware, only datum with a maximum of 400 input tokens were included. This removed at most 13.4\% of the data for each of the train/test splits of samsum. The xsum dataset was significantly larger and was further reduced to 15000 training and 3000 test examples. During PPO and at test for all models, the output was limited to 100 tokens, which is more than the maximum length of the ground-truth data in the training and test splits of both datasets. 

\subsection{Results} Table~\ref{tab:length_diffs}, focusing first on the first three columns, shows the average excess length of the output produced by the different models relative to the ground truth. Length is defined as the number of whitespace delineated entities. The baseline non-fine-tuned model was the most verbose. The maximum likelihood trained model produced shorter outputs, however, fine-tuning was more effective for samsum than xsum in this regard. The RL model produced output lengths very close to the ground truth for both datasets and outperformed the other models by a large margin.

Table~\ref{tab:scores} displays BLEURT and length-adjusted ROUGE scores as described above, as well as traditional ROUGE scores for reference. Maximum likelihood fine-tuning improved performance on both BLEURT and length adjusted ROUGE measures, and the RL model in turn outperformed the maximum likelihood optimized model. Qualitatively, the output of the RL model was consistently clean and concise. In contrast, the output of the maximum likelihood model contained repetition of sections of the input and prior output as well as occasional random tokens. These effects were present for both datasets but were more pronounced for samsum. In contrast, the output for xsum appeared to be significantly verbose even when accounting for the extra tokens due to repetition. 

As an additional test, performance of the baseline and maximum likelihood models was evaluated after light post-processing. The baseline models often started their response with a statement similar to ``Sure, here is a summary..." followed by a newline character even when instructed to output the summary only. For the cleaned version, the text leading up to the first newline character was removed if the response contained a newline and started with ``Sure,". The maximum likelihood models required two types of cleanup. First, repetition of the input had to be removed. Second, for xsum, the output was significantly verbose relative to the ground truth and had to be shortened. There is not a principled way to achieve the latter, but a reasonable attempt given the nature of the data (extreme summarization) is to throw away everything \emph{after} the first newline. This assumes that the beginning of the summary contains highly relevant information---whether or not this assumption is correct is an empirical question. For both datasets, truncating the output after the first newline character was also a reasonable heuristic to remove repetition\footnote{This is in contrast to separately attempting to remove repetition using a more complex decoding scheme~\citep[e.g.][]{su2022contrastive}, which may be necessary with other data.}. Thus, this simple transformation was uniformly applied to both datasets to achieve the dual aims. The results are displayed in the last two columns of Tables~\ref{tab:length_diffs} and~\ref{tab:scores}. Post-processing the outputs of the maximum likelihoods models bridged their performance gap with the RL models, with the cleaned up versions coming out slightly ahead on some measures. 

\input{table_length_diffs.tex}
\input{table_scores.tex}

\section{Discussion} Reinforcement learning-based fine-tuning produced cleaner outputs and beat maximum likelihood tuned models on semantic and n-gram based evaluation metrics. However, light post-processing of the maximum likelihood outputs was able to bridge the observed performance gap for the datasets analyzed here. The results of this work can be interpreted from two perspectives. First, from a practical perspective, the immediate utility of the method may initially appear diminished by the good performance of maximum likelihood with post-processing. However, the post-processing applied to the xsum dataset was to arbitrarily truncate the outputs, which is a procedure whose utility is idiosyncratic to specific types of data. Thus, it would still be useful to attempt the method across a range of other challenging fine-tuning scenarios.

Second, the current work provides a general framework that can further be built upon. An advantage of the RL-based approach is the ability to intuitively encode undesirable outputs via the reward model and to optimize the policy for each state to be cognizant of long-term cumulative reward. This was demonstrated with basic classes of outputs, but more significant gains may be realized by encoding more complex classes. For example, hallucinations and valid but poor summaries can be included, which could possibly be produced in an automated fashion. In order to train a reward model for these kinds of outputs, it may be necessary to modify the reward function from being token-based to being sequence-based, as in such cases it only makes sense to mark an output as ``bad'' after at least part of the sequence or even the entire sequence is produced. This would pose a greater challenge for policy optimization because the reward signal would be sparse and delayed in comparison to using token-based rewards and fitting the model may require greater care.

A disadvantage of the approach presented is the significant additional computational burden of fitting all of the reinforcement learning related models, especially running the LLM forward during policy optimization. While the focus here has been on last-mile fine-tuning to a specific dataset, if some additional generalization error is tolerable, it may be possible to incur these costs once by training on combined representative data.

The use of RL for last-mile fine-tuning was demonstrated for summarization, but the same approach can be used for any task. Finally, the goal of the current work was to contrast the performance of RL with maximum likelihood within an as identical as possible experimental setup rather than to produce state-of-the-art results. However, the procedure presented is independent of the size of the model, whether the model is quantized, and whether all of the model's parameters are fine-tuned or LoRA is employed, which all impact absolute performance. In addition, although the framing and focus has been on LLMs, the procedure can likely easily be modified to work with specialized state-of-the-art task-dependent architectures, 
and serve as a more general additional option when selecting among optimization algorithms for fine-tuning.

\vspace*{\fill}

\bibliographystyle{unsrtnat}
\bibliography{refs}

\end{document}

%% file: table_length_diffs.tex
\begin{table}[!h]
\centering
\caption{\label{tab:length_diffs}Extra length relative to ground truth}
\centering
\begin{tabular}[t]{lrrrrr}
\toprule
  & Base & Max LL & RL & Base cleaned & Max LL cleaned\\
\midrule
samsum & 39.2 & 12.7 & 2.9 & 28.9 & 5.8\\
xsum & 44.7 & 42.0 & -1.1 & 39.6 & -1.8\\
\bottomrule
\end{tabular}
\end{table}

%% file: table_scores.tex
\begin{table}[!h]
\centering
\caption{\label{tab:scores}Evaluation metrics}
\centering
\begin{tabular}[t]{l>{\raggedleft\arraybackslash}p{0.25in}>{\raggedleft\arraybackslash}p{0.25in}>{\raggedleft\arraybackslash}p{0.25in}>{\raggedleft\arraybackslash}p{0.25in}>{\raggedleft\arraybackslash}p{0.25in}>{\raggedleft\arraybackslash}p{0.25in}>{\raggedleft\arraybackslash}p{0.25in}>{\raggedleft\arraybackslash}p{0.25in}>{\raggedleft\arraybackslash}p{0.25in}>{\raggedleft\arraybackslash}p{0.25in}}
\toprule
\multicolumn{1}{c}{ } & \multicolumn{5}{c}{samsum} & \multicolumn{5}{c}{xsum} \\
\cmidrule(l{3pt}r{3pt}){2-6} \cmidrule(l{3pt}r{3pt}){7-11}
\begingroup\fontsize{8}{10}\selectfont  \endgroup & \begingroup\fontsize{8}{10}\selectfont Base\endgroup & \begingroup\fontsize{8}{10}\selectfont Max LL\endgroup & \begingroup\fontsize{8}{10}\selectfont RL\endgroup & \begingroup\fontsize{8}{10}\selectfont Base cleaned\endgroup & \begingroup\fontsize{8}{10}\selectfont Max LL cleaned\endgroup & \begingroup\fontsize{8}{10}\selectfont Base\endgroup & \begingroup\fontsize{8}{10}\selectfont Max LL\endgroup & \begingroup\fontsize{8}{10}\selectfont RL\endgroup & \begingroup\fontsize{8}{10}\selectfont Base cleaned\endgroup & \begingroup\fontsize{8}{10}\selectfont Max LL cleaned\endgroup\\
\midrule
bleurt & 0.51 & 0.57 & 0.58 & 0.54 & 0.58 & 0.32 & 0.37 & 0.48 & 0.33 & 0.49\\
\midrule
la-rouge1-F1 & 0.21 & 0.39 & 0.43 & 0.25 & 0.42 & 0.14 & 0.18 & 0.35 & 0.15 & 0.36\\
la-rouge1-precision & 0.21 & 0.39 & 0.44 & 0.25 & 0.43 & 0.14 & 0.18 & 0.35 & 0.15 & 0.36\\
la-rouge1-recall & 0.21 & 0.39 & 0.43 & 0.25 & 0.42 & 0.14 & 0.18 & 0.35 & 0.15 & 0.36\\
\midrule
la-rouge2-F1 & 0.08 & 0.20 & 0.21 & 0.10 & 0.22 & 0.03 & 0.07 & 0.15 & 0.03 & 0.15\\
la-rouge2-precision & 0.08 & 0.20 & 0.21 & 0.10 & 0.22 & 0.03 & 0.06 & 0.15 & 0.03 & 0.15\\
la-rouge2-recall & 0.08 & 0.20 & 0.21 & 0.10 & 0.22 & 0.03 & 0.07 & 0.15 & 0.03 & 0.15\\
\midrule
la-rougeL-F1 & 0.16 & 0.32 & 0.35 & 0.19 & 0.35 & 0.09 & 0.13 & 0.28 & 0.10 & 0.29\\
la-rougeL-precision & 0.16 & 0.32 & 0.36 & 0.20 & 0.35 & 0.09 & 0.13 & 0.28 & 0.10 & 0.29\\
la-rougeL-recall & 0.16 & 0.31 & 0.35 & 0.19 & 0.34 & 0.09 & 0.13 & 0.28 & 0.10 & 0.29\\
\midrule
rouge1-F1 & 0.31 & 0.46 & 0.50 & 0.35 & 0.50 & 0.21 & 0.26 & 0.39 & 0.22 & 0.40\\
rouge1-precision & 0.21 & 0.43 & 0.50 & 0.26 & 0.49 & 0.14 & 0.18 & 0.41 & 0.15 & 0.43\\
rouge1-recall & 0.65 & 0.60 & 0.56 & 0.63 & 0.58 & 0.45 & 0.54 & 0.39 & 0.44 & 0.39\\
\midrule
rouge2-F1 & 0.11 & 0.24 & 0.24 & 0.13 & 0.25 & 0.05 & 0.10 & 0.16 & 0.05 & 0.17\\
rouge2-precision & 0.08 & 0.22 & 0.25 & 0.10 & 0.25 & 0.03 & 0.07 & 0.17 & 0.03 & 0.18\\
rouge2-recall & 0.25 & 0.30 & 0.27 & 0.25 & 0.30 & 0.10 & 0.21 & 0.16 & 0.10 & 0.16\\
\midrule
rougeL-F1 & 0.23 & 0.38 & 0.41 & 0.27 & 0.40 & 0.14 & 0.19 & 0.31 & 0.15 & 0.33\\
rougeL-precision & 0.16 & 0.35 & 0.41 & 0.20 & 0.40 & 0.09 & 0.13 & 0.33 & 0.10 & 0.35\\
rougeL-recall & 0.50 & 0.49 & 0.46 & 0.49 & 0.48 & 0.31 & 0.40 & 0.31 & 0.30 & 0.32\\
\bottomrule
\end{tabular}
\end{table}